\documentclass{article}

\usepackage[preprint]{corl_2026} 
\usepackage{booktabs} 
\usepackage{enumitem}
\usepackage{graphicx}
\usepackage{wrapfig}
\usepackage{soul}
\usepackage{amsmath}
\usepackage{xspace}
\usepackage[most]{tcolorbox}
\usepackage{hyperref}

\tcbset{
    boxsep=1pt,
    top=3pt,
    bottom=3pt,
    left=2pt,
    right=2pt,
}

\newcommand{\ours}{MuSe\xspace}

\title{Multisensory Continual Learning: Adapting Pretrained Visuomotor Policies to Force}
\vspace{-0.2cm}
\vspace{-0.7cm}
\author{
  \textbf{Jaden Clark \quad Changhao Wang \quad Yihuai Gao \quad Seongheon Hong} \\
  \textbf{Hojung Choi \quad Mark Cutkosky \quad Yifan Hou \quad Shuran Song} \vspace{0.1cm} \\
  Stanford University
} \vspace{-0.5cm}

\begin{document}
\maketitle

\vspace{-8mm}
\begin{abstract}
Robot manipulation often depends on sensory data beyond vision, especially in contact-rich tasks where force, tactile, or audio feedback reveals interaction states not directly visible from images. Yet such modalities are hardware- and task-specific, and large multisensory datasets remain scarce, making it impractical to pretrain policies with every sensor they may encounter. We study \textit{multi-sensory continual learning}: adapting a pretrained robot policy to new tasks with newly introduced modalities while preserving performance under the original sensor suite. We propose MultiSensory World Model (\ours), which integrates limited multisensory data into pretrained vision-only policies using  multi-stage fusion, multisensory future prediction, and experience replay on pretraining data. We instantiate \ours by adding force-torque sensing to a pretrained vision-only policy and evaluate it on real-world manipulation tasks. Experiments show that \ours not only performs well on contact-rich finetuning tasks but also has performance gains on pretraining tasks - suggesting that a modest multisensory dataset can enhance general model capability beyond the finetuning distribution. Project website: \href{https://jadenvc.github.io/multisensory-continual-learning/}{https://jadenvc.github.io/multisensory-continual-learning/}
\end{abstract}
\vspace{-0.1cm}
\vspace{-0.1cm}


\vspace{-0.4cm}
\section{Introduction}
\vspace{-0.2cm}
\label{sec:intro}

Large-scale vision-action pretraining has become a powerful paradigm for robot learning, but vision alone does not fully capture the physical interaction state required for manipulation. Contact-rich tasks often depend on signals that are only indirectly visible, such as contact forces and slippage. Additional modalities such as force-torque (F/T), tactile, and audio sensing provide direct feedback about these hidden interaction dynamics, enabling robots to adapt their behavior between precise trajectory tracking and compliant contact.
In practice, however, such modalities are often hardware- and task-specific, and datasets that include them are far smaller than vision-only datasets. It is therefore impractical to include every possible sensor during large-scale pretraining. Robot policies must instead adapt post hoc, leveraging abundant visual pretraining while efficiently incorporating limited data from new sensors.

\begin{figure}[t]
    \centering
    \vspace{-3mm}
    \includegraphics[width=\linewidth]{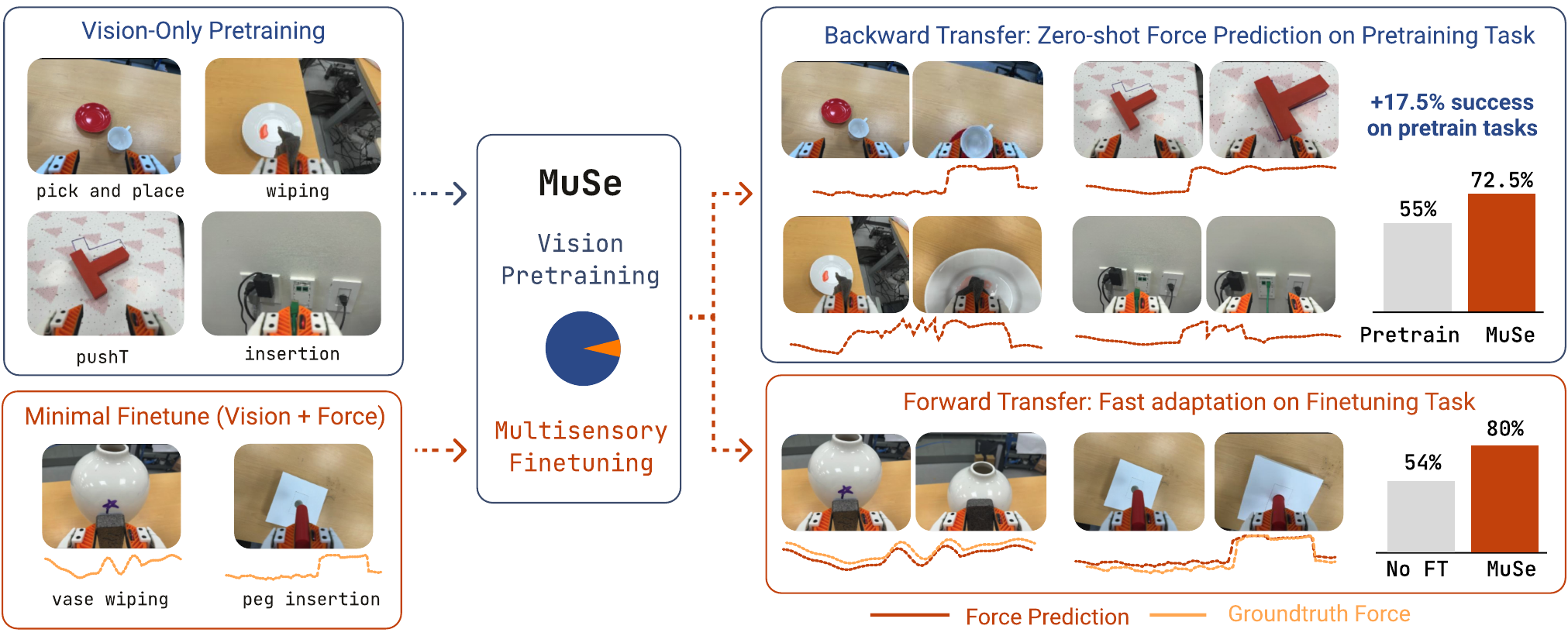}
    \vspace{-5mm}
    \caption{
    \textbf{Multisensory continual learning.}
    A policy is first pretrained on diverse vision-action data without force-torque (F/T) labels, then adapted with a small amount of multisensory data from new contact-rich tasks. 
    \ours enables \textit{\textbf{improved performance on pretraining tasks with no additional task-specific data}} (backward transfer), zero-shot F/T prediction where no F/T supervision was collected (cross-modal generalization), and improved performance on finetuning tasks that benefit from F/T sensing (forward transfer). 
    Orange curves show predicted F/T signals, light-orange show ground-truth F/T, and bar plots compare \ours task success rate with the corresponding baselines.
    }
    \vspace{-0.2cm}
    \label{fig:teaser}
\end{figure}
\vspace{-0.1cm}
We formalize this challenge as \textbf{multi-sensory continual learning}: extending a pretrained policy to incorporate new sensory modalities for new tasks, while retaining performance on the original tasks.  Prior work has largely studied multi-sensory learning and continual learning in isolation: \textit{multi-sensory learning} typically assumes access to all modalities from the start of training, whereas \textit{continual learning} focuses on acquiring new tasks under a fixed sensory input. In contrast, we study their intersection: augmenting an existing policy with new modalities for new tasks, while maintaining performance on the original tasks.

Concretely, a successful multi-sensory continual learning method should aim for the following goals: 
\vspace{-2mm}
\begin{itemize}[leftmargin=3mm] 
\vspace{-1mm}    \item \textbf{Forward transfer:} Improve performance on new tasks using the new modality, compared with training from scratch.
    
\vspace{-1mm}    \item \textbf{Backward transfer:} Retain or improve the performance on pretraining tasks after learning new tasks with new sensory data.
    
\vspace{-1mm}    \item \textbf{Cross-modal generalization:} Understand and utilize a newly introduced modality on tasks where it was never observed. 
\end{itemize}

Although this setting better reflects practical deployment, it also introduces several technical challenges. Naively finetuning on new tasks with new sensor modality often leads to catastrophic forgetting, degrading performance on pretraining tasks. Additionally, incorporating new modalities requires the model to reason over new input dimensions that were absent during pretraining, without either over-relying on or ignoring them. This requires an algorithm that preserves prior knowledge acquired during pretraining while properly incorporating new input modalities through an effective cross-modal representation. 

To address these challenges, we propose \ul{Mu}lti-\ul{Se}nsory World Model (\textbf{\ours}), a general framework for multisensory continual learning that enables efficient integration of new sensory modalities into pretrained policies (Fig. \ref{fig:teaser}). Concretely, we study how to augment a pretrained vision-action model with force-torque (F/T) input for unseen tasks. We identify the following components as critical to its success:

\vspace{-1mm}
\begin{itemize}[leftmargin=3mm]
\vspace{-1mm}
    \item \textbf{Multi-stage fusion:} \ours fuses F/T with vision and proprioception immediately after encoding by embedding all modalities into a shared token space. The resulting tokens are processed jointly by the transformer, enabling full cross-modal attention and stronger downstream prediction than late modality-specific adapters alone, as commonly used in prior work.
\vspace{-1mm}
    \item \textbf{Multi-sensory future prediction (i.e., world modeling):} \ours is trained to predict future visual observations, F/T observations, and actions. This objective encourages a unified representation across modalities that generalizes beyond the finetuning data and improves backward transfer.

\vspace{-1mm}
    \item \textbf{Experience Replay:} Naively fine-tuning models on new tasks with multisensory data leads to degradation in pretraining task performance. We find that a simple strategy of co-fine-tuning on pretraining data, with F/T inputs masked when unavailable, enables the model to retain performance on pretraining tasks while learning new contact-rich behaviors.
\end{itemize}
\vspace{-3mm}

Through \ours, we find that F/T sensing is surprisingly general: even when learned from limited contact-rich data, it can be predicted and exploited across a broad range of robot manipulation tasks. Conditioning on F/T enables richer representation learning for contact-rich behavior, while predicting future F/T provides an estimate of the compliance profile required for precise forceful interaction \cite{hou2025adaptive}.

In summary, our key contributions are:
\vspace{-1mm}

\begin{itemize}[leftmargin=3mm]

    \vspace{-1mm}\item We formalize multi-sensory continual learning for robotics and introduce a comprehensive evaluation protocol.
    \vspace{-1mm}\item We introduce \ours, a framework for multi-sensory continual learning that identifies key design decisions for strong performance, including the training objective, network architecture, and training procedure: shared future prediction, multi-stage fusion, and experience replay.
    \vspace{-1mm}\item We demonstrate the effectiveness of our approach through extensive real-world experiments on both contact-rich and pick-and-place tasks, showing improvements in real-world success rate and offline cross-modal generalization.
\end{itemize}
\vspace{-1mm}

\vspace{-0.2cm}
\section{Related Works}
\vspace{-0.1cm}
\label{sec:related_works}
\paragraph{Multisensory robot learning.}
Recent work has extended robot policies with touch, force, and audio, showing that these modalities provide contact state, slip, and interaction dynamics that are difficult to infer from vision alone \cite{choi2026wild,heng2025vitacformer,liu2024maniwav,zhang2023efficient,zhu2025touch,yin2026osmo,jones2025fuse,zheng2026omnivta,yuan2026vtam}. Visuo-tactile world models and video-tactile-action models further demonstrate the value of explicitly modeling contact dynamics for physical interaction \cite{zheng2026omnivta,yuan2026vtam}. However, multisensory datasets remain far smaller and less diverse than vision-only pretraining corpora. We therefore study a sensor-incremental setting: learning from limited multisensory data while retaining the broad capabilities acquired during visual pretraining.
\vspace{-0.2cm}
\paragraph{Continual learning in robotics.}
Continual learning addresses catastrophic forgetting when robot policies are adapted to new data \cite{thrun1995lifelong,lesort2020continual,liu2023libero}. Recent work on pretrained vision-language-action policies shows that replay, regularization, and guidance can preserve prior performance or grounding during post-training \cite{liu2026pretrained,huang2026delock}, while related work adapts general vision-language models for robotic actions and evaluates generalization in generalist robot policies \cite{zitkovich2023rt2,driess2025knowledge,hancock2026actions,gao2025stargen}. These methods largely assume a fixed observation space. In contrast, our setting changes the sensor suite itself, requiring the policy to incorporate a new input modality while preserving behavior supported by the original modalities. Our approach follows the spirit of replay and co-training, but adds modality masking and fusion mechanisms for sensor expansion.
\vspace{-0.3cm}
\paragraph{Multisensory adaptation.}
The closest prior work is FuSe, which also extends generalist robot policies to heterogeneous sensor inputs through language grounding \cite{jones2025fuse}. Our setting differs in two ways: we target contact-rich manipulation, and we explicitly study how the new modality can transfer backward to the original vision-only tasks. CR-DAgger is also related in its use of force feedback for contact-rich manipulation, but it treats force as a residual correction for an existing policy, whereas we study expanding a generalist policy to new sensors~\cite{xu2025compliant}.
\vspace{-0.3cm}


\section{Method}
\vspace{-2mm}

\subsection{Problem Statement}
\vspace{-1mm}
We consider a pretrained robot model trained on trajectories comprising $n$ observation modalities and robot actions. Let $\mathbf{o_t}$  $= \{o_t^1, \ldots, o_t^n\}$ denote the observations at time $t$, where each $o_t^i$ corresponds to one modality, such as RGB images or proprioception. Given a history horizon $h$, the pretrained model takes as input a sequence of past observations and actions, $
    \{o_{t-h+1}, \ldots, o_t\}, 
    \{a_{t-h}, \ldots, a_{t-1}\},
$
and predicts future actions and observations over a prediction horizon $H$:
$
    \{o_{t+1}, \ldots, o_{t+H}\},
    \{a_t, \ldots, a_{t+H-1}\}, 
$

The goal of \ours is to adapt this pretrained model to incorporate an additional sensor modality, denoted $o_t^{n+1}$, such as force-torque. During fine-tuning, we assume access to a smaller multisensory dataset $\mathcal{D}_{\text{new}}$ collected on new tasks that includes the new modality, as well as access to the original pretraining data $\mathcal{D}_{\text{pre}}$, which may not contain $o^{n+1}$. The objective is to learn a model that can condition on the new modality when it is available, predict its future values when supervision is provided, and preserve the capabilities learned from the original pretraining distribution.

This setting differs from standard multimodal learning, where all modalities are assumed to be available throughout training, and from standard continual learning, where the sensor suite is fixed. Instead, we study \textit{multi-sensory continual learning}: adding a new sensor to a pretrained policy while preserving performance on the original tasks and modalities.

\begin{wrapfigure}{r}{0.6\textwidth}
    \centering
    \includegraphics[width=\linewidth]{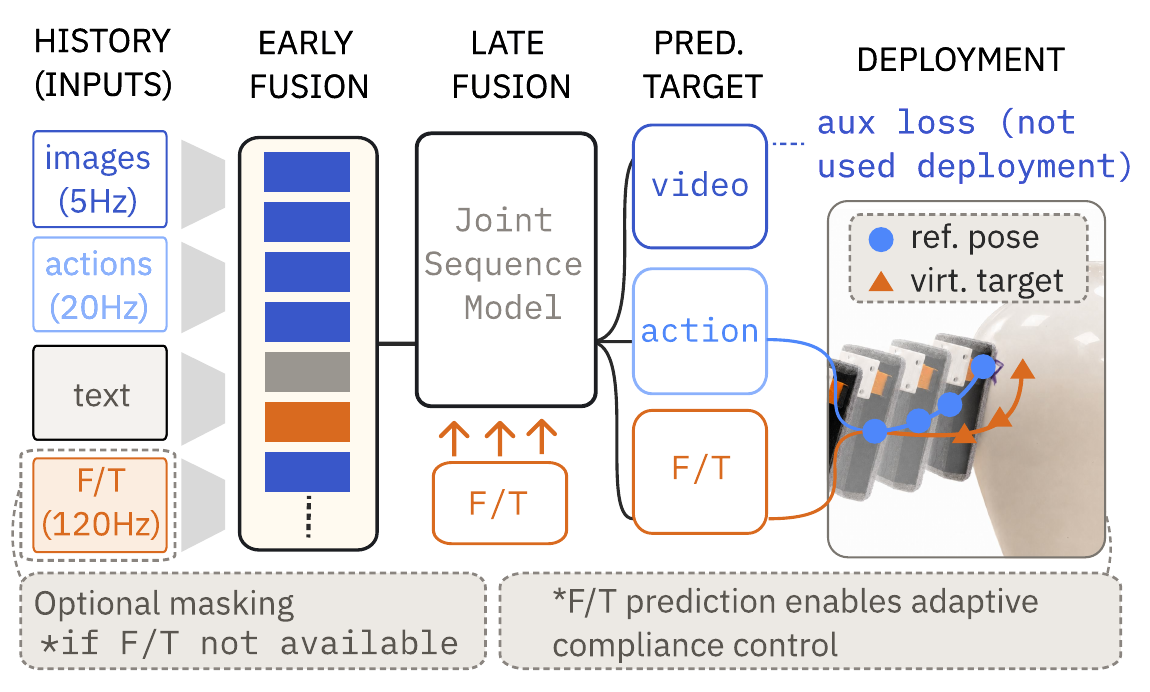}
    \caption{
    \textbf{\ours architecture.}
    \ours encodes image, proprioceptive, language, and optional force-torque (F/T) histories with modality-specific encoders, then fuses them through token-level early fusion and late fusion via cross-attention adapters. 
    The joint sequence model predicts future actions, F/T signals, and auxiliary video frames, with unavailable F/T inputs and losses masked during training. 
    At deployment, action predictions drive the policy to commanded reference pose, while predicted future F/T is used to set a virtual target for adaptive compliance control.
    }
    \vspace{-0.3cm}
    \label{fig:method}
     
\end{wrapfigure}

\vspace{0.6cm}

\subsection{Multi-stage Fusion}
\vspace{-0.3cm}

A central challenge in multisensory learning is how to combine modalities expressively without corrupting the pretrained representation. The new modality should interact deeply with vision, proprioception, language, and action, but it should not overwrite visual-action knowledge or cause the policy to over-attend to contact signals when they are uninformative. Existing multisensory robot learning methods typically address this  with a single fusion stage, either combining modalities early to maximize cross-modal interaction or injecting them later to preserve the pretrained representation ~\cite{heng2025vitacformer,choi2026wild,zhu2025touch,jones2025fuse,xu2025compliant}. MuSe goes beyond these single-stage designs by combining both pathways (Fig. \ref{fig:method}).

\ours extends the pretrained model by adding a modality-specific encoder and projection layer for the newly introduced modality $o_t^{n+1}$. The projection maps the new modality features into the shared embedding space used by the pretrained observation and action tokens. The resulting tokens are concatenated with the existing observation, proprioception, and action tokens along the sequence dimension before being processed by the backbone model. This early-fusion pathway places all modalities in a shared token space, allowing the backbone to perform cross-modal attention throughout its layers. In contrast to approaches that condition the model only through late FiLM layers, adapters, or cross-attention modules, early fusion gives the new modality direct access to the same sequence-processing pathway as the original modalities.

We further use late fusion as a complementary pathway. Features from the new modality are injected into intermediate transformer layers through lightweight cross-attention adapters. Together, the early- and late-fusion pathways form a multi-stage fusion architecture: early fusion supports global token-level interaction, while late fusion strengthens modality-specific conditioning throughout the backbone.

\vspace{-0.4cm}

\subsection{Multisensory Future Prediction}
\vspace{-0.2cm}

Multi-stage fusion enables the policy to condition on an added modality, but conditioning alone does not ensure that the new sensor improves the shared representation. To encourage the model to integrate new sensory information into a unified, physically grounded latent space, we train \ours with a flexible multisensory future prediction objective. In addition to predicting future actions, the model predicts future original observations, such as video, as well as the newly introduced modality. Predicting the new modality encourages the representation to encode information from the added sensor, while predicting future images anchors this representation to visual dynamics that are shared across datasets.

Let $H$ denote the prediction horizon. Given a context ending at time $t$, \ours is trained to predict future actions and future observations across all available modalities. We separate the observation prediction objective into two parts: prediction of the original modalities $o^{1:n}$ and prediction of the newly introduced modality $o^{n+1}$. The resulting objective is
\[
    \mathcal{L}
    =
    \lambda_a \mathcal{L}_{\mathrm{act}}
    +
    \lambda_o \mathcal{L}_{\mathrm{obs}}
    +
    \lambda_{n+1} \mathcal{L}_{\mathrm{new}}.
\]
Here, $\mathcal{L}_{\mathrm{act}}$ supervises prediction of actions over the horizon $a_{t:t+H-1}$, $\mathcal{L}_{\mathrm{obs}}$ supervises prediction of the original observation modalities $o^{1:n}_{t+1:t+H}$, and $\mathcal{L}_{\mathrm{new}}$ supervises prediction of the newly introduced modality $o^{n+1}_{t+1:t+H}$. Each term is applied only when the corresponding target is available, allowing the same objective to support datasets with different modality coverage.

\vspace{-0.3cm}
\subsection{Experience Replay}
\vspace{-0.2cm}
Naively fine-tuning the pretrained model only on the new multisensory data can lead to catastrophic forgetting: the model may adapt to the new sensor and task distribution while losing performance on the original pretraining tasks. To mitigate this, we use \textit{experience replay} during finetuning. Each training batch is sampled from a mixture of the new multisensory dataset $\mathcal{D}_{\text{new}}$ and the original pretraining dataset $\mathcal{D}_{\text{pre}}$.

For samples from $\mathcal{D}_{\text{new}}$, the model receives the expanded multisensory input when available and is supervised with the relevant action, original-observation, and new-modality prediction losses. For samples from $\mathcal{D}_{\text{pre}}$, the new modality is absent. We replace the missing modality input with a learned modality-mask token and mask out $\mathcal{L}_{\mathrm{new}}$, while retaining supervision for actions and original observation prediction. Thus, the same multisensory prediction objective can be applied across both datasets using per-modality availability masks.

\vspace{-0.2cm}
\subsection{Implementation}
\vspace{-0.1cm}
We instantiate \ours by augmenting the Unified Video-Action (UVA) model~\cite{li2025unified} with force-torque (F/T) sensing. UVA jointly models video and action sequences with a decoupled action decoder for efficient policy inference. To incorporate F/T, we add the two fusion pathways described above: early fusion encodes F/T histories with a causal convolutional encoder and prepends the resulting tokens to the pretrained token sequence, while late fusion injects F/T features through cross-attention adapters. We also add an F/T diffusion head following the UVA action head design.

At each decision step, the policy conditions on the past four image frames, past 16 actions, and synchronized F/T histories from both fingers. It predicts 16 future actions, four future image frames, and 96 future F/T samples per finger. Each F/T target consists of six high-rate 6D wrench samples per action step.

The predicted future wrench trajectory also provides an interface for adaptive compliance control~\cite{hou2025adaptive}. At deployment, \ours predicts expected wrenches in the robot tool frame, which can be converted into a virtual force target for a 6D task-space admittance controller. Following Adaptive Compliance Policy (ACP)~\cite{hou2025adaptive}, the predicted F/T trajectory is used to reconstruct the stiffness matrix and virtual target for the low-level controller, allowing the robot to remain stiff in free space and adapt compliance when contact or constraints are anticipated.


\section{\ours with Force-Torque Sensing: A Case Study}
\label{sec:result}
\begin{figure}[h]
    \centering
    \includegraphics[width=\linewidth]{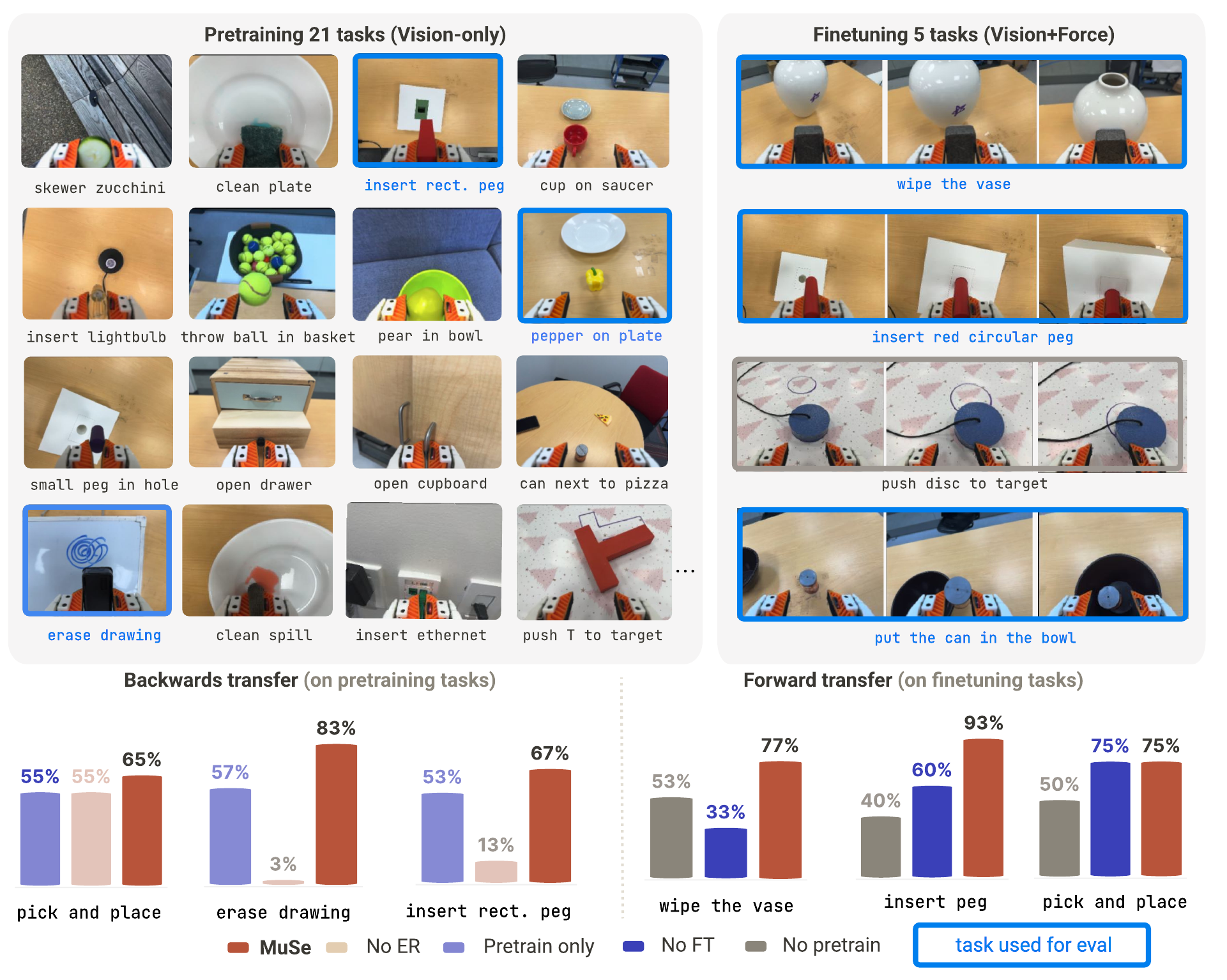}
    \vspace{-5mm}
    \caption{
    We pretrain a policy (with no F/T) on 21 tasks, then finetune on 5 \textit{unique tasks} containing F/T information that do not exist in pretraining. We then evaluate \ours on 3 sets of pretraining tasks to measure backwards transfer, and 3 sets of finetuning tasks for forward transfer. \ours outperforms the pretrained model on all three pretraining tasks despite adding no additional data on these tasks, in addition to outperforming models with no pretraining and no F/T for forward transfer. 
    }
    \label{fig:experiments}
    \vspace{-5mm}
\end{figure}
\vspace{-0.2cm}

\subsection{Cross-Modal Generalization}
\label{{sec:cross_modal_generalization}}
\begin{wrapfigure}{r}{0.6\textwidth}
    \centering
    \vspace{-0.7cm}
    \includegraphics[width=\linewidth]{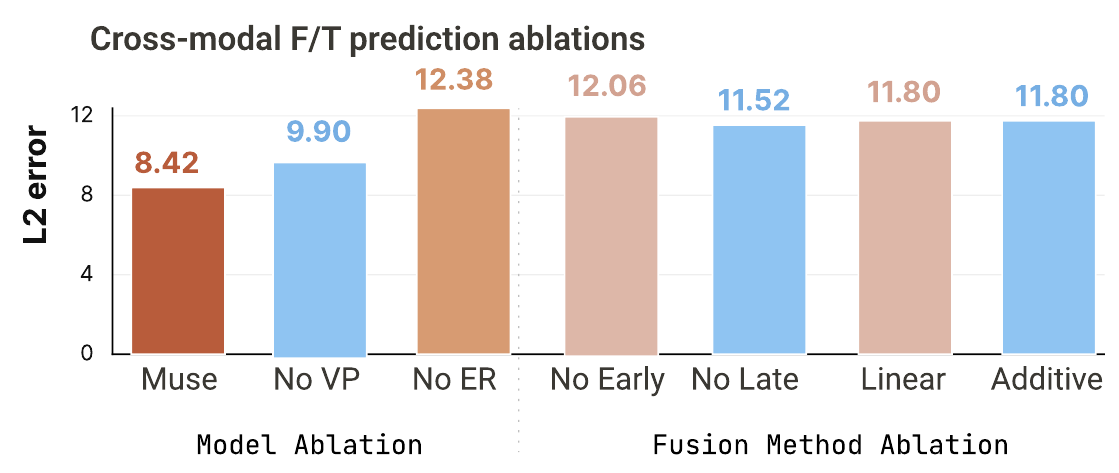}
    \caption{
     L2 error of predicted F/T signals on pretraining tasks.
    \ours achieves the lowest error, while weakening fusion, removing video prediction, or removing experience replay increases prediction error.
    }
    \vspace{-0.2cm}
    \label{fig:cross_modal_generalization}
\end{wrapfigure}

\textbf{Experiment Setup.} 
We collect demonstrations using UMI-FT~\cite{choi2026wild}, which equips each gripper finger with a CoinFT sensor and streams all modalities to the control desktop for real-time inference. We then pretrain the UVA backbone on 1,271 episodes from 21 tasks using only visual observations and actions. We then extend it with force--torque (F/T) inputs and finetune on 434 episodes from five contact-rich tasks with F/T supervision. Although F/T signals are recorded during data collection, they are withheld during vision-only pretraining. This protocol allows us to evaluate whether force-aware behavior and F/T prediction can transfer to tasks where force supervision was never provided.

We evaluate \ours across three settings: cross-modal generalization of force--torque prediction to tasks where F/T supervision was never provided, forward transfer to new multimodal contact-rich tasks, and backward transfer to the original pretraining tasks after multimodal finetuning (Fig. \ref{fig:experiments}).
\vspace{-0.2cm}

We first evaluate whether the model can predict F/T signals on pretraining tasks where F/T was recorded but never used for supervision. This tests whether multimodal finetuning learns a transferable contact representation rather than only fitting force signals on the finetuning tasks. We report the F/T prediction error as the mean Euclidean error over each 6D force-torque vector. Specifically, for each predicted timestep and each finger, we compute the L2 distance between the predicted and measured 6D F/T vector, then average this error over all timesteps, chunks, left and right fingers, and evaluation trajectories.

We ablate two design choices: the F/T fusion mechanism and the training objectives. For fusion, we compare our full model against variants that remove early F/T tokens, remove late cross-attention adapters, or replace token concatenation with linear or additive fusion. For training, we remove video-action prediction objectives either throughout training or only during finetuning, and we also evaluate the effect of removing experience replay.

\begin{figure}[h]
    \centering
    \vspace{-3mm}
    \includegraphics[width=\linewidth]{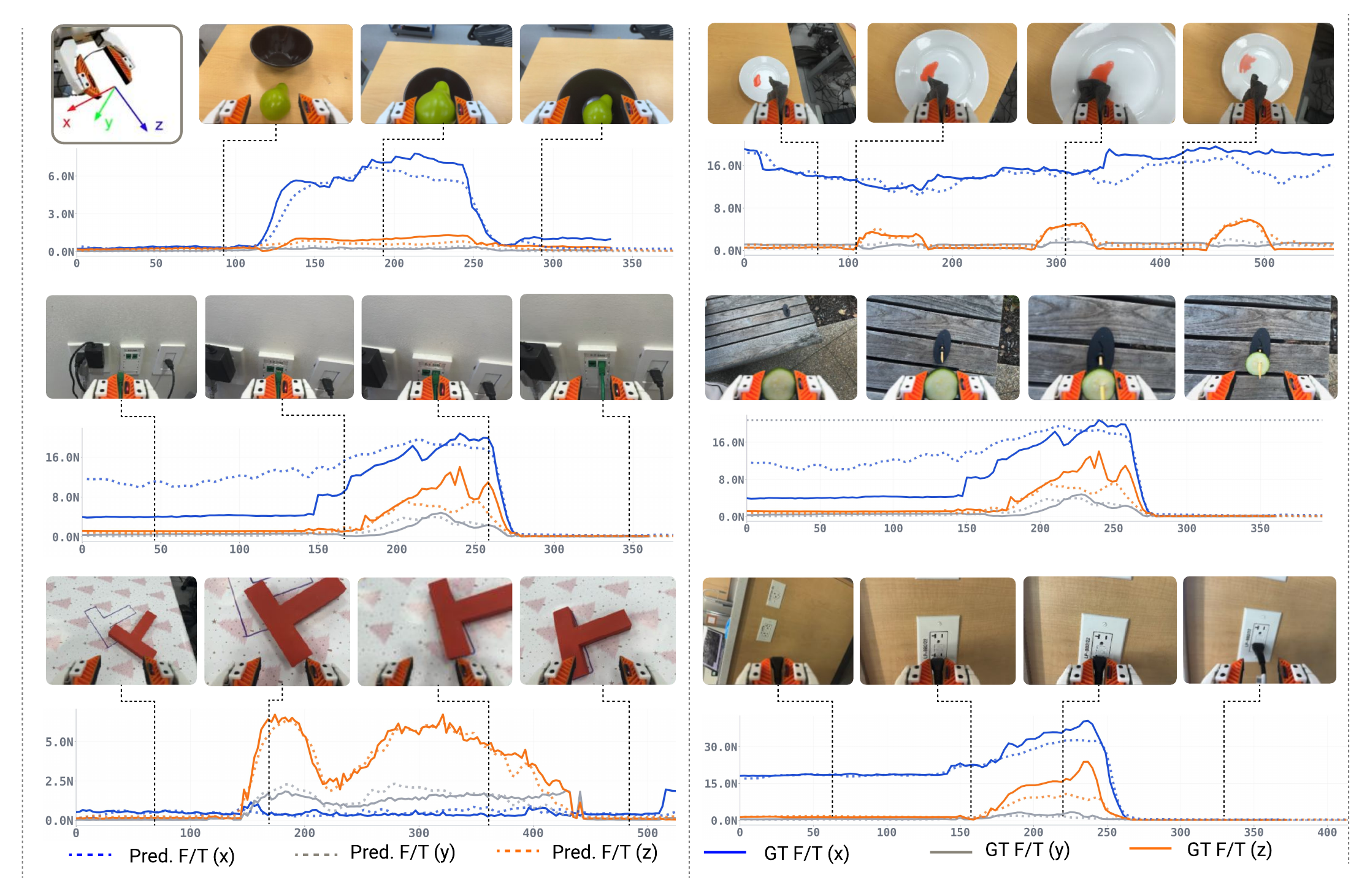}
    \vspace{-5mm}
    \caption{
    Cross-modal generalization of F/T prediction on pretraining tasks where F/T signals were recorded but never used for supervision. \ours accurately predicts changes in F/T across all three axes (defined in top left). Failure modes include under or overestimation (middle left, bottom right).
    }
    \label{fig:crossmodal}
\end{figure}

\ours obtains the lowest F/T prediction error, with a L2 error of 8.424. Weakening the fusion mechanism consistently hurts performance: late-only fusion reaches 12.061, while \textbf{Linear Fusion} and \textbf{Additive Fusion} reach 11.518 and 11.800. Removing the video prediction auxiliary loss also increases error, with \textbf{No VP} reaching 9.90. The largest degradation comes from removing experience replay: \textbf{No ER} reaches 12.376, the highest error among all variants (Fig. \ref{fig:cross_modal_generalization}). These results indicate that transferable F/T prediction benefits from both early and late fusion, video-action prediction, and continued exposure to the pretraining distribution during multimodal finetuning (Fig. \ref{fig:crossmodal}).

\begin{tcolorbox}[colback=blue!10]
\textbf{Takeaway:} Early and late fusion, video prediction, and experience replay are each key in learning transferable F/T representations.
\end{tcolorbox}

\vspace{-0.4cm}
\subsection{Forward Transfer}

We then evaluate whether multimodal finetuning enables a vision-action pretrained policy to acquire force-aware behaviors from a small number of demonstrations. We compare \ours against two baselines: \textbf{No F/T}, which removes F/T inputs and the F/T prediction loss while keeping the same pretrained UVA backbone and finetuning data, and \textbf{No Pretrain}, which trains the same multisensory architecture from scratch on the finetuning demonstrations containing F/T.

We evaluate on three downstream tasks. \textbf{Vase wiping} requires the robot to erase drawings from a curved vase while regulating contact force; evaluation varies drawing location, color, and eraser size, including orange drawings unseen during finetuning. \textbf{Peg insertion} requires inserting red pegs into high-friction corrugated holes, where jams must be detected and corrected through contact feedback. \textbf{Pick and place} evaluates whether adding F/T preserves performance on tasks where force feedback is less central, with perturbed object locations and distractors. Notably, the model is evaluated on pick and place tasks that exist in the finetuning dataset but not in pretraining.

\ours improves most on the contact-rich tasks. It achieves 11.5/15 success on \textbf{vase wiping}, compared to 5/15 for \textbf{No F/T} and 8/15 for \textbf{No Pretrain}, and 13/15 on \textbf{peg insertion}, compared to 9/15 and 6/15, respectively. On \textbf{pick and place}, both methods with pretraining achieve 7.5/10, while \textbf{No Pretrain} struggles to handle distractors and reaches 5/10. This suggests that F/T finetuning does not degrade performance when force feedback is not the main bottleneck, and that \ours is able to leverage pretraining to enhance model generalization. The failures of \textbf{No F/T} are primarily contact-regulation errors: it either applies too much or too little force during wiping, and it cannot reliably recover from jams during insertion. In contrast, \ours uses predicted F/T to regulate contact while retaining strong pick-and-place performance.

\begin{tcolorbox}[colback=blue!10]
\textbf{Takeaway:} \ours leverages F/T to improve success rate on contact-rich tasks,  while vision-action pretraining improves generalization on task variations.
\end{tcolorbox}

\vspace{-2mm}
\subsection{Backward Transfer}
\vspace{-2mm}
Finally, we evaluate whether multimodal finetuning preserves the original vision-only skills while transferring the newly learned F/T modality back to the pretraining tasks. We compare against \textbf{No ER}, which removes experience replay from the original pretraining data during finetuning, and \textbf{Pretraining Only}, which uses the original vision-only UVA policy without multimodal finetuning.

We evaluate on three pretraining tasks. \textbf{Whiteboard wiping} requires the robot to erase a drawing, either with the eraser already grasped or after first picking it up. Success is scored as 0.5 if more than 50\% of the drawing is removed and 1.0 if more than 90\% is removed. \textbf{Peg insertion} uses peg and hole geometries unseen in finetuning data, including smaller cylindrical pegs and rectangular pegs. \textbf{Pick and place} includes pretraining-only object-goal combinations such as peppers on plates, peppers in pots, and grapes in pots, with distractors and perturbed object locations.

\ours achieves 12.5/15 success on \textbf{whiteboard wiping}, 10/15 on \textbf{peg insertion}, and 6.5/10 on \textbf{pick and place}, outperforming both baselines. \textbf{Pretraining Only} reaches 8.5/15, 8/15, and 5.5/10, respectively, indicating that vision-only pretraining retains useful task knowledge but lacks force-aware execution. \textbf{No ER} drops to 0.5/15 on \textbf{whiteboard wiping} and 2/15 on \textbf{peg insertion}, showing severe forgetting without recall. Qualitatively, \textbf{Pretraining Only} often attempts the correct behavior but fails to regulate contact. \textbf{No ER} often forgets the original task structure or overfits to finetuning-specific strategies (e.g. twisting peg into hole - which works for circular pegs but not for rectangular pegs). These results suggest that experience replay is important both for preserving visual skills and for transferring F/T prediction back to earlier tasks.

\begin{tcolorbox}[colback=blue!10]
\textbf{Takeaway:} Experience replay prevents forgetting, and F/T input and output can improve—not just preserve—performance on pretraining tasks.
\end{tcolorbox}
\vspace{-1mm}
\vspace{-0.2cm}
\section{Conclusion}
\label{sec:conclusion}
\vspace{-0.2cm}
\vspace{-1mm}
We propose \ours, a general framework for adapting pretrained policies to new modalities through multisensory future prediction, multi-stage fusion, and episodic recall. We instantiate this framework with force--torque sensing as a case study on top of a vision-action policy. Across real-world experiments, \ours learns transferable F/T representations, improves forward transfer to contact-rich tasks, and preserves or improves performance on the original vision-only task distribution. These results suggest that pretrained robot policies can be expanded with new sensors after pretraining, without requiring large-scale multisensory data from the start.
\label{sec:limitation}

\noindent \textbf{Limitations and future works:} We study multi-sensory continual learning through a single sensor-expansion case: adding force--torque sensing to a pretrained vision-action policy. Future work will evaluate whether the same principles extend to other modalities, such as tactile, audio, depth, or multi-sensor combinations. Our experiments also rely on experience replay with access to the original pretraining data; developing methods that make the approach easier to deploy when prior data cannot be stored will be another future direction. Finally, our real-world evaluation focuses on a limited set of tabletop manipulation tasks. Scaling to more diverse robots, longer-horizon tasks, and deformable objects remains an important direction for future work.

\clearpage
\acknowledgments{Jaden Clark is supported by the Knight-Hennessy Fellowship and the NSF Graduate Research Fellowships Program (GRFP). This work was supported in part by the NSF Award \#2143601, \#2037101, and \#2132519, Toyota Research Institute, Amazon, Stanford System-X, Stanford HAI Center. We would like to thank Mandi Zhao and Maximilian Du for presentation feedback as well as all members of the REAL lab at Stanford for their detailed feedback on paper drafts and experiment directions. The views and conclusions contained herein are those of the authors and should not be interpreted as necessarily representing the official policies, either expressed or implied, of the sponsors.}


\bibliography{example}  
\newpage
\section{Appendix}
\vspace{-0.2cm}
\label{sec:appendix}
Project website: \href{https://multisensorylearning.github.io/multisensory-world-model/}{MultisensoryLearning}

\subsection{Performance with Fixed Compliance}

MuSe uses force--torque (F/T) information in two ways during deployment: the policy conditions on measured F/T history, and the predicted future F/T trajectory is used to set an adaptive compliance profile for the low-level controller. To disentangle these two effects, we evaluate MuSe on contact-rich finetuning tasks with adaptive compliance disabled. In this setting, the model still uses measured F/T history as input, but its predicted F/T trajectory is not used to modulate the compliance controller. We compare this variant against the full MuSe model, a \textbf{No F/T} baseline that removes F/T input and F/T prediction, and a \textbf{No Pretrain} baseline trained from scratch on the multisensory finetuning data (which conditions on and predicts F/T).

\begin{table}[h]    \centering    \caption{    \textbf{Effect of adaptive compliance on contact-rich finetuning tasks.}    We report real-world success over 15 trials. \textbf{No F/T} removes force input and force prediction, \textbf{No Pretrain} trains the same multisensory architecture from scratch, and \textbf{MuSe w/o Compliance} deploys MuSe without using predicted F/T for adaptive compliance. Best results are bolded.    }    \label{tab:no_compliance}    \begin{tabular}{lcccc}        \toprule        Task & No F/T & No Pretrain & MuSe w/o Compliance & MuSe \\        \midrule        Vase wiping & 5/15 & 8/15 & 8/15 & \textbf{11.5/15} \\        Peg insertion & 9/15 & 6/15 & 11/15 & \textbf{13/15} \\        \bottomrule    \end{tabular}\end{table}

Full MuSe performs best on both tasks, indicating that F/T prediction is useful not only as an auxiliary representation-learning objective, but also for adapting stiffness. On vase wiping, disabling adaptive compliance reduces performance from 11.5/15 to 8/15. This task is particularly sensitive to compliance because the robot must maintain contact with a curved surface while applying enough force to erase the drawing without exceeding the F/T safety limit. Many failures without adaptive compliance were contact-regulation failures: the robot either applied insufficient force and failed to erase the mark, lost contact with the vase surface, or exceeded the force limit and terminated the rollout. Peg insertion also shows a meaningful drop when adaptive compliance is removed, from 13/15 to 11/15. In this task, F/T input is useful to sense when the peg first contacts the hole, infer the direction and angle of the corrugated hole, detect when the peg is stuck, and decide how to correct its motion. However, compliance remains useful because the robot must also become less stiff near contact to search for the hole, avoid hard impacts, and stop small jams before the peg is fully wedged. Overall, these results suggest that MuSe benefits from both forms of force usage. Conditioning on F/T history improves contact-state estimation and recovery behavior, while predicted F/T enables adaptive compliance that improves contact regulation and prevents dangerous forces during contact.

\subsection{Diffusion Policy Baseline}
\label{app:diffusion_policy_baseline}

We additionally evaluate a Diffusion Policy (DP) baseline to ablate the role of
\emph{multi-sensory future prediction}, one of the central design choices in
MuSe. MuSe is trained as a world model that predicts future visual observations,
future F/T observations, and future actions. This objective is intended to
encourage a shared representation across vision, force, proprioception, and
action, so that F/T prediction can generalize beyond the small multisensory
finetuning dataset. In contrast, a standard diffusion policy directly models
robot actions and does not predict future visual observations. Thus, DP provides
a policy-only counterpart to MuSe that uses the same data and replay protocol,
but removes future visual prediction from the training objective.

During pretraining, we train DP on the full vision-only pretraining dataset using
image histories and robot proprioception as input, and future robot actions as
the prediction target. Since the pretraining data do not provide F/T supervision,
the pretrained DP model is a vision-proprioception-action policy. We then
finetune DP using the same multisensory continual-learning pipeline as MuSe. We
add an F/T encoder so that the model can condition on force histories, and we add
an F/T prediction head so that the finetuned model predicts both future actions
and future F/T signals. For samples from the multisensory finetuning dataset, DP
receives image, proprioception, and F/T histories as input and is supervised with
both action and F/T prediction losses. For replayed samples from the pretraining
dataset, where F/T is unavailable, we replace the F/T input with a mask token and
mask out the F/T prediction loss while retaining the action loss. 

For cross-modal generalization, we report the F/T prediction error as the mean Euclidean error over each 6D force-torque vector. Specifically, for each predicted timestep and each finger, we compute the L2 distance between the predicted and measured 6D F/T vector, then average this error over all timesteps, chunks, left and right fingers, and evaluation trajectories. DP achieves low error on the finetuning tasks, with an
average F/T prediction error of \(6.04\). However, its error
increases substantially on pretraining tasks, reaching an average of
\(18.27\). These tasks were included during vision-action
pretraining and replay, but they never provided F/T supervision. The large gap
between finetuning and pretraining tasks suggests that DP can fit the supervised
F/T signals in the finetuning distribution, but does not acquire a force
representation that transfers broadly to the pretraining distribution.

\begin{table}[!htbp]
    \centering
    \small
    \caption{
    Diffusion Policy baseline as an ablation of multi-sensory future prediction.
    Both methods use experience replay and masked F/T supervision when F/T labels
    are unavailable. Lower is better.
    }
    \label{tab:dp_vs_muse_future_prediction}
    \begin{tabular}{lcccc}
        \toprule
        Method & Act. pred. & F/T pred. & Img. pred. & F/T error $\downarrow$ \\
        \midrule
        Diffusion Policy & Yes & Yes & No  & 18.27 \\
        MuSe             & Yes & Yes & Yes & 8.42  \\
        \bottomrule
    \end{tabular}
\end{table}

Table~\ref{tab:dp_vs_muse_future_prediction} summarizes the comparison between
DP and MuSe. Both methods use the same multisensory finetuning protocol,
including experience replay, masked F/T inputs for replayed pretraining samples,
and masked F/T losses when force supervision is unavailable. The key difference
is that MuSe additionally predicts future visual observations. The substantially
lower pretraining-task F/T error of MuSe suggests that future image prediction and late fusion adapters
help ground force prediction in visual dynamics shared across tasks, leading to
stronger cross-modal generalization beyond the finetuning distribution.


Overall, this baseline strengthens the ablation in
Section 4.1: removing video prediction within
MuSe increases F/T prediction error, and replacing MuSe with a policy-only
diffusion baseline without future visual prediction or late fusion leads to an even larger
degradation on pretraining tasks. These results suggest that multi-sensory
future prediction is important not only for fitting F/T on the finetuning tasks,
but also for learning a transferable cross-modal representation that supports
backward transfer.

We additionally evaluate diffusion policy performance in the real world. We find that it achieves strong performance on finetuning tasks, but significantly degrades in performance on contact-rich backward transfer evaluation - suggesting that modeling future images and our \ours architecture are key to learning effective cross-modal representations. Results are displayed in Tables \ref{tab:real_world_forward} and \ref{tab:real_world_backward}.

\begin{table}[t]
\centering
\caption{Real-world task success on forward transfer (success credit / trials).}
\label{tab:real_world_forward}
\begin{tabular}{lccc}
\hline
Method & Wiping & Peg & Pick and Place \\
\hline
No F/T & 5/15 & 9/15 & 7.5/10 \\
No Pretrain & 8/15 & 6/15 & 5/10 \\
Diffusion Policy & 10/15 & 10/15 & 6/10 \\
MuSe & \textbf{11.5/15} & \textbf{13/15} & \textbf{7.5/10} \\
\hline
\end{tabular}
\end{table}

\begin{table}[t]
\centering
\caption{Real-world task success on backward transfer (success credit / trials).}
\label{tab:real_world_backward}
\begin{tabular}{lccc}
\hline
Method & Wiping & Peg & Pick and Place \\
\hline
Pretraining Only & 8.5/15 & 8/15 & 5.5/10 \\
No ER & 0.5/15 & 2/15 & 5.5/10 \\
Diffusion Policy & 6.5/15 & 5/15 & 6/10 \\
MuSe & \textbf{12.5/15} & \textbf{10/15} & \textbf{6.5/10} \\
\hline
\end{tabular}
\end{table}

\subsection{Training Details}
\label{app:training_details}

\paragraph{Model architecture.}
Our model is built on the Masked Autoregressive (MAR) framework~\cite{li2024autoregressive}
using the MAR-Large backbone with 479M parameters. The backbone consists of a
16-layer transformer encoder and a 16-layer transformer decoder, each with
embedding dimension 1024 and 16 attention heads. Images are tokenized using a
frozen KL-16 VAE with stride 16, mapping each $256{\times}256$ frame to
$16{\times}16=256$ spatial tokens of dimension 16. With patch size 1, each token
embedding has dimension 16 and is projected to 1024 dimensions by a learned
linear layer. We condition on 4 history frames and predict 4 future frames,
yielding $4 \times 256 = 1024$ visual tokens per video sequence.

\paragraph{Token construction.}
The encoder receives a single 1024-token sequence formed by combining projected
embeddings from all conditioning modalities. History video tokens are projected
to the shared 1024-dimensional embedding space. Proprioception is represented by
concatenating end-effector position, 6D rotation, gripper width, and 6D rotation
relative to the episode start, resulting in a 16-dimensional vector per timestep.
This vector is projected to 1024 dimensions and broadcast to match the visual
token sequence length. Action tokens follow the same pattern with an independent
projection. All tokens receive a combined positional embedding, formed by adding
a learned temporal embedding over output frames and a learned spatial embedding
over image patch positions. During training, classifier-free guidance is applied
to language conditioning by randomly dropping language embeddings with
probability 0.1.

\paragraph{Force--torque encoding.}
Force--torque (F/T) measurements from the left and right finger sensors are
encoded independently by a causal convolutional encoder. We use the raw F/T
representation, where each proprioceptive timestep contains 6 high-frequency
F/T samples, and each sample contains 6 wrench values
$(F_x, F_y, F_z, T_x, T_y, T_z)$ in Newtons and Newton-meters. This gives a
36-dimensional F/T vector per action timestep. For each of the 4 history frames,
the corresponding 4 proprioceptive timesteps are concatenated into a
$4 \times 36 = 144$-dimensional vector and passed through a three-layer MLP
$(144 \to 256 \to 512 \to 1024)$, producing one embedding per frame. This yields
a $[B,4,1024]$ tensor for each sensor, which is then broadcast to 1024 tokens.

We compare several strategies for fusing the resulting left and right F/T
embeddings into the main token sequence. In the \textit{linear} early-fusion
variant, F/T embeddings are concatenated channel-wise with all other modality
tokens and jointly projected through a single linear layer, increasing the input
dimension by $2 \times 1024$. In the \textit{additive} early-fusion variant, F/T
embeddings are added to the fused token features through zero-initialized linear
projections, preserving the pretrained model at initialization. In the
\textit{token prepend} variant, F/T embeddings are prepended as $2 \times 4$
explicit sequence tokens with learned positional embeddings. In the
\textit{FiLM} late-fusion variant, a mean-pooled global F/T vector modulates
intermediate encoder features through Feature-wise Linear Modulation adapters
inserted after the last $d$ transformer blocks, with adapter weights
zero-initialized. In the \textit{cross-attention} late-fusion variant, the full
left-and-right F/T token sequence attends into encoder features through
cross-attention adapters after the last $d$ blocks, gated by a scalar parameter
initialized to zero.

\paragraph{Training data.}
Table~\ref{tab:training_data} summarizes the training data used for vision-only
pretraining and multisensory finetuning. Finetuning with \ours uses the full dataset (all pretrain and all finetune data) with uniform weights.

\begin{table}[t]
\centering
\small
\begin{tabular}{llr}
\toprule
Split & Language label & Episodes \\
\midrule
\multicolumn{3}{l}{\textbf{Pretraining data}} \\
& erase the drawing & 278 \\
& pick up the beans and place them in the bowl & 94 \\
& pick up the pear and place it in the bowl & 140 \\
& place the block in the bowl & 99 \\
& place the rectangular peg in the hole & 50 \\
& place the white peg in the red hole & 49 \\
& place the green rectangular peg in the hole & 43 \\
& place the pizza in the bin & 22 \\
& place the small purple peg in the hole & 50 \\
& pick up the red cup and place it on the white saucer & 43 \\
& place the pepper in the pot & 50 \\
& throw the object into the bin & 18 \\
& wipe the surface with the rag & 20 \\
& open the cupboard & 20 \\
& wipe the surface with the sponge & 20 \\
& insert the ethernet cable into the port & 18 \\
& insert the connector into the socket & 18 \\
& push the T-block to the target & 22 \\
& open the drawer & 17 \\
& skewer the zucchini & 100 \\
& insert the lightbulb & 100 \\
\midrule
\multicolumn{2}{r}{\textbf{Pretraining total}} & \textbf{1271} \\
\midrule
\multicolumn{3}{l}{\textbf{Finetuning data}} \\
& place the can in the bowl & 60 \\
& push the disc to the target & 20 \\
& erase the drawing on the vase & 160 \\
& place the peg in the hole & 144 \\
& pick up the white cup and place it on the red saucer & 50 \\
\midrule
\multicolumn{2}{r}{\textbf{Finetuning total}} & \textbf{434} \\
\bottomrule
\end{tabular}
\caption{
Training data used for vision-only pretraining and multisensory finetuning.
Rows are grouped by language label, with finetuning datasets sharing the same
language label combined into a single entry.
}
\label{tab:training_data}
\end{table}

\paragraph{Training modes.}
We train the model under multiple task modes, sampled uniformly at random in
each batch. This allows a single model to support video generation, forward
dynamics prediction, inverse dynamics prediction, and action inference.

In \textit{video model mode}, the model predicts future video frames from
history frames alone using a masked autoregressive video diffusion loss
$\mathcal{L}_{\mathrm{video}}$. In \textit{dynamics model mode}, ground-truth
future actions are additionally provided as conditioning, and the same video
prediction loss is applied to train a forward dynamics model. In \textit{full
dynamics model mode}, both the video and action heads are active, jointly
optimizing $\mathcal{L}_{\mathrm{video}} + \mathcal{L}_{\mathrm{act}}$. In
\textit{policy model mode}, future visual tokens are replaced by learned fake
latents, so the model predicts actions purely from history and is supervised by
the action diffusion loss $\mathcal{L}_{\mathrm{act}}$. In \textit{inverse model
mode}, both history frames and observed future frames are provided to the
encoder, and the model predicts the intervening actions. Our full model is
trained with all five modes, while the no-video-prediction baseline is trained
with policy mode only.

\paragraph{Output heads.}
Both the video and action output heads use diffusion-based decoders with depth 6
and width 1024. They are trained with 1000 diffusion steps and evaluated with
100 denoising steps. The F/T prediction head uses the same architecture and
diffusion schedule, and predicts the concatenated left and right wrench
trajectory $[B,T,72]$ in the raw F/T space. For episodes without F/T sensors,
F/T loss terms are masked out using a per-episode availability flag.

\paragraph{Optimization.}
All models are finetuned from the MAR-Large ImageNet checkpoint. We train on
8 NVIDIA RTX Pro 6000 GPUs with a total batch size of 40, using AdamW with
learning rate $1 \times 10^{-4}$, weight decay 0.02, and momentum parameters
$(0.9,0.95)$. We train until validation loss convergence. An exponential moving
average (EMA) of the model weights is maintained throughout training and used
for evaluation.

\subsection{Deployment Details}
\textbf{Hardware:} We deploy all policies on the same UMI-FT setup used for data collection. The robot consists of a UR5e arm equipped with a WSG50 parallel gripper and UMI-FT fingers, with one CoinFT six-axis force--torque sensor mounted on each finger. The end-effector also includes the same wrist-mounted iPhone used in the UMI-FT platform, providing  RGB observations for policy inference. During robot deployment, RGB observations, robot proprioception, action history, and left/right finger force--torque measurements are streamed to the control desktop and synchronized online.

\textbf{Inference:} The deployed policy uses the same temporal interface as training. RGB frames are provided at 5 Hz, actions and proprioception at 20 Hz, and force--torque measurements at 120 Hz. At each decision step, \ours conditions on the past four image frames, past 16 actions, robot proprioception, and synchronized force--torque histories from both fingers. It predicts a 16-step horizon of future end-effector actions and future force--torque values. Each action step is associated with six force--torque predictions, matching the 120 Hz force stream relative to the 20 Hz action stream.

The action output is executed as a sequence of end-effector reference poses. In addition, \ours uses its predicted future force--torque trajectory to drive adaptive compliance. For each future action step, we aggregate the corresponding high-rate force--torque predictions by taking the maximum predicted wrench magnitude over the six force samples. We then apply a moving-average filter across the predicted horizon and use the filtered force profile to compute a virtual target for the task-space admittance controller. This allows the robot to remain stiff during free-space motion while becoming compliant when contact is anticipated.

\textbf{Controllers:} Following the UMI-FT/ACP control stack, measured finger wrenches are transformed into the robot tool frame and combined to provide external wrench feedback for task-space admittance control. The tool center point is set near the center of the two fingertips, which improves alignment during contact-rich behaviors such as wiping and insertion. Before each rollout, the CoinFT sensors are tared to remove bias. For safety, we terminate rollouts when the measured force exceeds 20 N or the measured torque exceeds 6 Nm.

\begin{figure}[h]
    \centering
    \includegraphics[width=\linewidth]{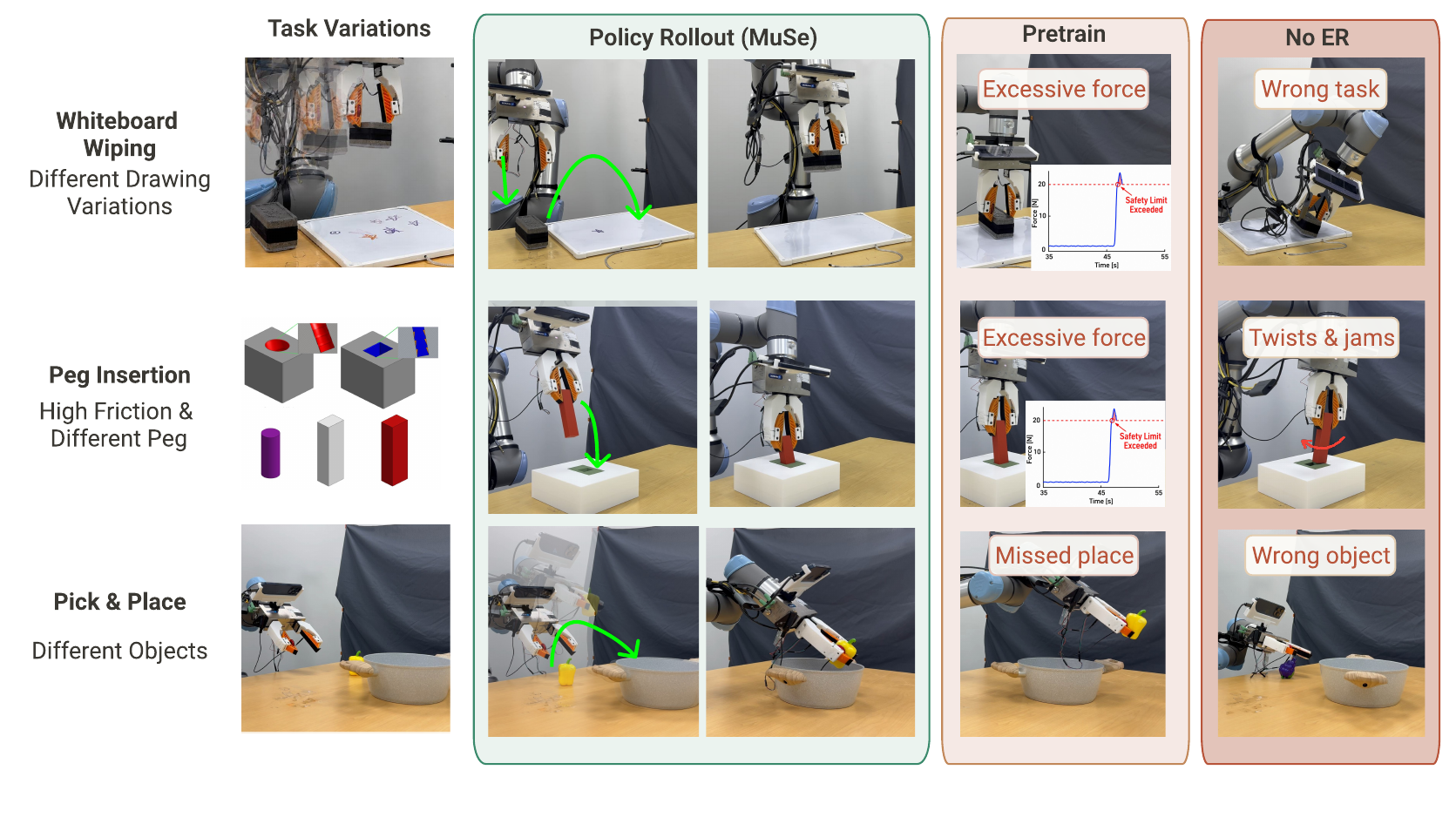}
    \vspace{-5mm}
    \caption{
    \textbf{Backward Transfer}: Evaluation details for pretraining tasks. Left column shows initial task variation during evaluation, second and third columns show successful rollouts with \ours, fourth column shows failure modes with no finetuning (typically wrong application of force), and fifth column shows failure modes of No ER model (typically wrong task strategy).
    }
    \label{fig:b}
\end{figure}

\subsection{Evaluation Details}

\paragraph{Pretraining-task evaluation.} We evaluate backward transfer on three task families from the original vision-only pretraining distribution. These tasks test whether multisensory finetuning preserves the pretrained policy's broad manipulation skills while enabling the newly learned force--torque representation to improve execution. \textbf{Whiteboard wiping.} The robot must erase a drawing from a whiteboard, either with the eraser already grasped or by first picking up the eraser from the table. This task tests both object acquisition and precise contact regulation: the robot must grasp the eraser reliably, reach the drawing, and apply enough normal force to erase the mark without exceeding the F/T safety limit. Applying too little force leaves the drawing unerased, while applying too much force triggers the safety stop. We evaluate 15 trials total: 10 trials begin with the eraser already in the gripper, while 5 trials require the robot to first pick up the eraser. Across trials, we perturb the drawing color, drawing location, board pose, and eraser initial pose. A rollout receives full credit if more than 90\% of the drawing is removed, 0.5 credit if more than 50\% is removed, and zero credit otherwise. Common failures include failing to grasp the eraser, missing the board, applying insufficient force, or exceeding the F/T safety limit. 

\textbf{Peg insertion (diverse).} The robot must insert pegs into high-friction, serrated holes. This task evaluates whether the policy can use force feedback to detect contact, recover from jams, and adapt its insertion strategy across different peg and hole geometries. Unlike the finetuning peg-insertion task, which uses a single red flat-grip peg and a white hole, the pretraining-task evaluation uses geometrically and visually diverse peg-hole pairs. We evaluate 15 trials total: 7 trials use a red rectangular peg in a green hole, 3 trials use a red rectangular peg in a white hole, and 5 trials use a purple peg in a white hole. The task is difficult without force feedback because friction and serration make the peg likely to jam, and the robot must actively escape these jams rather than simply slide the peg into place. A rollout receives full credit only if the peg is inserted to the bottom of the hole. Failures include missing the hole, jamming, exceeding the F/T safety limit, or getting stuck before insertion. We observe that the vision-only pretrained policy frequently becomes stuck without necessarily exceeding the F/T limit, causing the policy to predict near-zero actions and drift out of distribution. The No ER model often uses a finetuning-specific insertion strategy: for rectangular pegs, it attempts to twist the peg to escape jams or lock it into the hole, a behavior that can work for circular pegs seen during finetuning but fails for rectangular pegs unless the peg is precisely aligned. In contrast, MuSe can observe increasing F/T during jams and adjust its motion to wriggle out before continuing insertion. 

\textbf{Pick and place (diverse).} The robot must complete pretraining-only pick-and-place tasks. We evaluate 10 trials total: 4 trials place a pepper in a pot, 2 trials place grapes in a pot, and 4 trials place a pepper on a plate. Object positions are perturbed, and distractors are added to evaluate whether the policy preserves visual generalization and task selection after multisensory finetuning. Unlike wiping and insertion, this task does not primarily require force feedback; instead, it tests whether adding F/T and finetuning on contact-rich tasks degrades performance on manipulation behaviors where vision and semantic grounding are more important. A rollout receives 0.5 credit if the robot successfully grasps the target object and full credit if it places the object in the correct goal region. Failures include selecting the wrong object, failing to grasp the target, dropping the object, or placing it outside the goal region. Both MuSe and the pretrained baseline have moderate success rates on this task, likely due to the limited amount of data collected for these pretraining pick-and-place variants.

\begin{figure}[h]
    \centering
    \includegraphics[width=\linewidth]{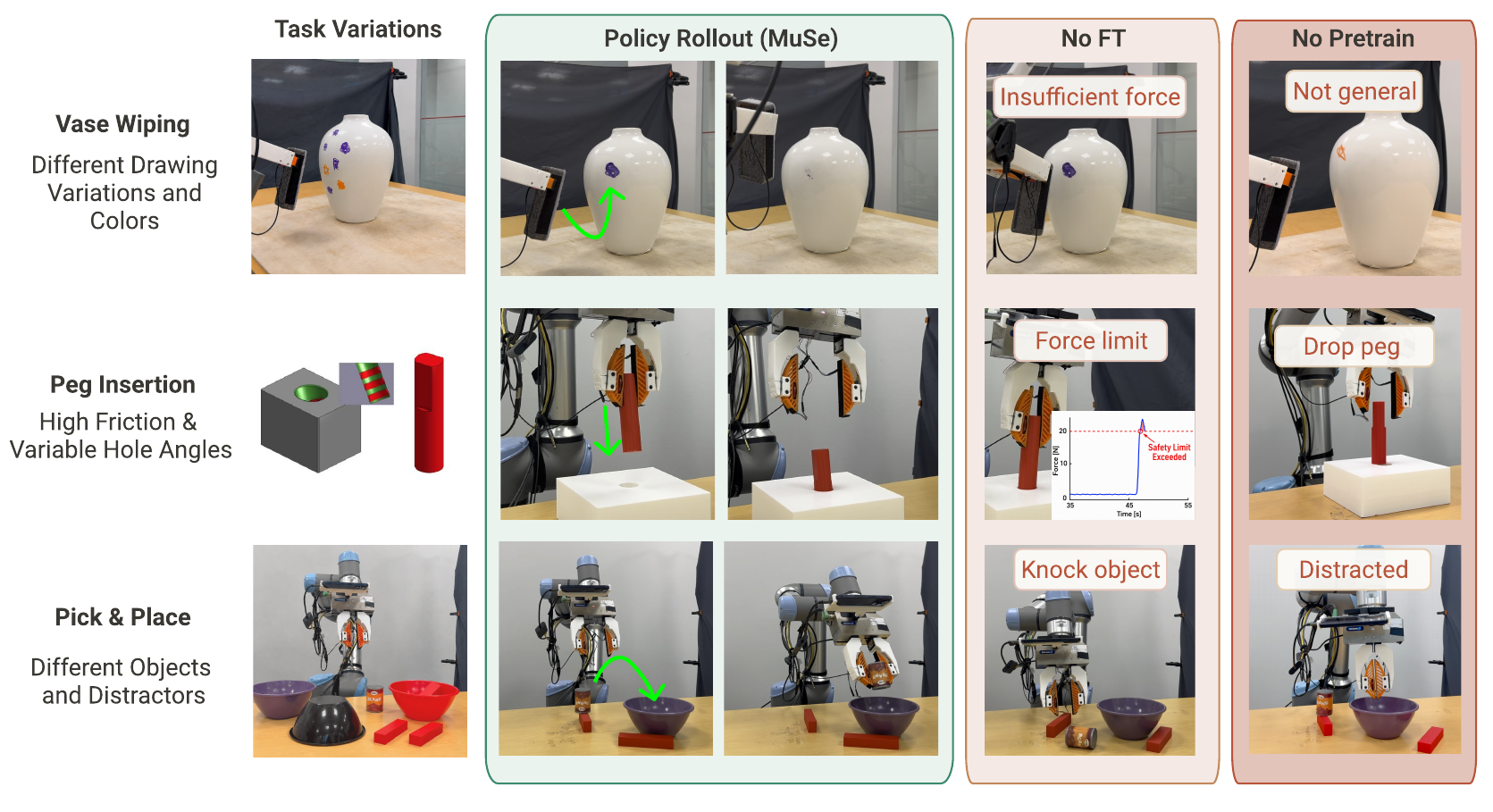}
    \vspace{-5mm}
    \caption{
    \textbf{Forward Transfer}: Evaluation details for finetuning tasks. Left column shows initial task variation during evaluation, second and third columns show successful rollouts with \ours, fourth column show failure modes with no F/T input (typically hits force limit or not enough application of force), and fifth column shows failure modes of No Pretraining model (typically failure to generalize).
    }
    \label{fig:f}
\end{figure}

\paragraph{Finetuning-task evaluation.} We evaluate forward transfer on three task families from the multisensory finetuning distribution. These tasks test whether a vision-action pretrained policy can efficiently acquire new contact-rich behaviors from limited force--torque data. 
\textbf{Vase wiping.} The robot starts with the eraser already grasped and must erase drawings from a curved vase. This task is related to whiteboard wiping but is more difficult because the robot must follow a curved surface while maintaining appropriate contact force. Too little force fails to erase the drawing, while too much force risks damaging the vase or exceeding the robot's F/T safety limit. We evaluate 15 trials total, varying drawing location, drawing color, and eraser size. The evaluation includes 4 orange drawings and 11 purple drawings. During finetuning, all vase drawings are purple, so orange drawings test whether the model can reuse visual wiping knowledge from pretraining. We also vary eraser size, with 3 trials using a large eraser and 12 trials using the normal eraser. A rollout receives full credit if more than 90\% of the drawing is removed, 0.5 credit if more than 50\% is removed, and zero credit otherwise. Failures include insufficient contact force, excessive force, missing the drawing, losing contact with the curved surface, or exceeding the F/T safety limit. 

\textbf{Peg insertion (narrow).} The robot must insert a red peg into a white high-friction corrugated hole. This task evaluates whether the model can use F/T input to detect contact, identify jams, and change insertion direction during precise insertion. Unlike the pretraining peg-insertion evaluation, which includes diverse peg colors, hole colors, and peg geometries, the finetuning task uses a single red flat-grip peg and a white hole. This makes the finetuning task narrower in visual and geometric diversity, but still contact-rich because the corrugated hole induces frequent jams. We evaluate 15 trials total. Across trials, the hole angle is varied between 20 degrees forward tilt and 20 degrees backward tilt. This perturbation is visually difficult to observe, requiring the policy to infer the insertion direction from force feedback during contact. A rollout receives full credit only if the peg is fully inserted to the bottom of the hole. Failures include missing the hole, jamming, exceeding the F/T safety limit, or failing to reach the bottom of the hole. Successful policies must respond to increasing F/T during contact and adjust compliance or motion before the peg becomes irrecoverably stuck. 

\textbf{Pick and place (narrow).} The robot must place cups on saucers and cans in bowls while object locations are perturbed and distractors are added. We evaluate 10 trials total: 3 trials place a cup on a saucer, and 7 trials place a can in a bowl. For the can-in-bowl task, we add red distractors and large objects to perturb the policy and test robustness to clutter. This task evaluates whether multisensory finetuning preserves performance on tasks where force feedback is less essential, while still allowing the model to use the same deployment interface as contact-rich tasks. These finetuning pick-and-place tasks use different objects and receptacles from the pretraining pick-and-place evaluation, allowing us to test whether finetuning preserves generic pick-and-place behavior rather than memorizing a shared object set. A rollout receives 0.5 credit if the robot successfully grasps the target object and full credit if it places the object in the correct goal region. Failures include picking the wrong object, failing to grasp the target, dropping the object, or placing it outside the goal region.

\end{document}